\documentclass[a4paper]{article}

\usepackage{INTERSPEECH2018}
\usepackage{slashbox}
\usepackage{graphicx}
\usepackage{wrapfig}
\usepackage[caption=false]{subfig}
\usepackage{multirow}
\usepackage{pbox}
\usepackage{booktabs}
\usepackage{xcolor}
\usepackage{soul}
\usepackage{color}
\usepackage[toc,page]{appendix}
\usepackage{hyperref}

\let\svthefootnote\thefootnote

\title{A Multi-Discriminator CycleGAN for Unsupervised Non-Parallel Speech Domain Adaptation}
\name{Ehsan Hosseini-Asl, Yingbo Zhou, Caiming Xiong, Richard Socher}

\address{
  Salesforce Research}
\email{\{ehosseiniasl,yingbo.zhou,cxiong,rsocher\}@salesforce.com}

\begin{document}

\maketitle
\begin{abstract}
Domain adaptation plays an important role for speech recognition models, in particular, for domains that have low resources.
We propose a novel generative model based on cyclic-consistent generative adversarial network (CycleGAN) for unsupervised non-parallel speech domain adaptation.
The proposed model employs multiple independent discriminators on the power spectrogram, each in charge of different frequency bands.
As a result we have 1) better discriminators that focus on fine-grained details of the frequency features, and 2) a generator that is capable of generating more realistic domain-adapted spectrogram.
We demonstrate the effectiveness of our method on speech recognition with gender adaptation, where the model only has access to supervised data from one gender during training, but is evaluated on the other at test time. 
Our model is able to achieve an average of $7.41\%$ on phoneme error rate, and $11.10\%$ word error rate relative performance improvement as compared to the baseline, on TIMIT and WSJ dataset, respectively. 
Qualitatively, our model also generates more natural sounding speech, when conditioned on data from the other domain.
  %Index terms should be included as shown below.
\end{abstract}

\let\thefootnote\relax\footnotetext{Sound demos can be found at~\href{https://einstein.ai/research/a-multi-discriminator-cyclegan-for-unsupervised-non-parallel-speech-domain-adaptation}{https://einstein.ai/research/a-multi-discriminator-cyclegan-for-unsupervised-non-parallel-speech-domain-adaptation}}
\addtocounter{footnote}{0}\let\thefootnote\svthefootnote

\section{Introduction}

Neural-based acoustic models have shown promising improvements in building automatic speech recognition (ASR) systems~\cite{Sak2014LongSM,Sainath2015DeepCN,Peddinti2015ATD,Senior2015AcousticMW}. 
However, it tends to perform poorly when evaluated on out-of-domain data, because of mismatch between the training and testing distribution (Table~\ref{tab:wsj-asr}). 

Domain mismatch is mainly due to variation in non-linguistic features, such as different speaker identity, unseen environmental noise, large accent variations, etc. 
Therefore, training a robust ASR system is highly dependent on factorizing linguistic features (text) from non-related variations, or adapting the inter-domain variations of source and target. 

Voice conversion (VC) has been widely used to adapt the non-linguistic variations, such as statistical methods~\cite{Stylianou1998ContinuousPT,Toda2007VoiceCB,Helander2010VoiceCU}, and Neural-based models~\cite{Chen2014VoiceCU,Nakashika2014VoiceCB,Desai2010SpectralMU,Mohammadi2014VoiceCU,Nakashika2014HighorderSM,Sun2015VoiceCU,Kaneko2017SequencetoSequenceVC}. 
However, traditional VC methods require parallel data of source and target that is difficult to obtain in practice.
In addition, the requirement of parallel data prevent these methods from using more abundant unsupervised data. Therefore, an unsupervised domain adaptation is desirable for building a robust ASR system. 

% In this paper, we propose a new generative model based on CycleGAN~\cite{CycleGAN2017} for unsupervised non-parallel domain adaptation. 
% Since differences in magnitude of frequency is the main variation across domains for spectrogram representations, we proposed to use multiple frequency-adaptable discriminators, that allows the model to learn the spectro-temporal variations during CycleGAN training. This will allow the generator to learn the mapping function which can convert spectrogram from source to target domain. It is shown that the original CycleGAN model is failing to learn the correct mapping function between domains, even with discriminator pretraining, and the generator collapses into learning an identity mapping function. 

In this paper, we propose a new generative model based on CycleGAN~\cite{CycleGAN2017} for unsupervised non-parallel domain adaptation. 
Since differences in magnitude of frequency is the main variation across domains for spectrogram representations, it is imperative that CycleGAN correctly catch the spectro-temporal variations between different frequency bands across domains during training. 
This will allow the generator to learn the mapping function which can convert spectrogram from source to target domain. 
In this paper, we show that the original CycleGAN model is failing to learn the correct mapping function between domains, and the generator collapses into learning an identity mapping function, which results in generating a noisy and unnatural-sounding audio.
%YZ: the "it is shown", do you actually mean "In this paper we show that"?
%YZ: robotic-sounding audio or unatrual-sounding audio ?

% catch we proposed to use multiple frequency-adaptable discriminators, that allows the model to learn the spectro-temporal variations during CycleGAN training. 

% This will allow the generator to learn the correct mapping function, that catch the temporal frequency differences between domains.

% \setlength{\tabcolsep}{4pt}
\renewcommand{\arraystretch}{0.65}
\begin{table}[tbp!]
\caption{ASR prediction mismatch when train/test on different genders, and when adapting using Multi-Discriminator CycleGAN, on WSJ (eval92) dataset}
  \label{tab:wsj-asr}
  \vspace{-1em}
\centering
\begin{tabular}{p{0.65cm}p{0.85cm}p{5.4cm}}
% \toprule
%  &  \\
%  &  \\
\toprule
 &  & \scriptsize Train on Male  \\
\midrule
\multirow{6}{1cm}{\scriptsize Test on Female} & \tiny True & \tiny \textcolor{blue}{CIBA} AGREED TO REMEDY THE \textcolor{blue}{OVERSIGHT}  \\
& \tiny Female & \tiny \textcolor{red}{SEVEN} AGREED TO REMEDY THE \textcolor{red}{OVER SITE} \\
& \tiny Female$\rightarrow$Male & \tiny \textcolor{blue}{CIBA} AGREED TO REMEDY THE \textcolor{red}{OVER SITE} \\
\cmidrule{2-3}
& \tiny True & \tiny A LITTLE \ldots NEWS COULD \textcolor{blue}{SOFTEN} THE \textcolor{blue}{MARKET'S} RESISTANCE \\
& \tiny Female & \tiny A LITTLE \ldots NEWS COULD \textcolor{red}{SOUTH IN} THE \textcolor{red}{MARKETS} RESISTANCE \\
& \tiny Female$\rightarrow$Male & \tiny A LITTLE \ldots NEWS COULD \textcolor{blue}{SOFTEN} THE \textcolor{blue}{MARKET'S} RESISTANCE \\
\midrule
\midrule
& \tiny  & \scriptsize Train on Female  \\
\midrule
 \multirow{6}{1cm}{\scriptsize Test on Male} & \tiny True & \tiny \textcolor{blue}{THEY EXPECT} COMPANIES TO \textcolor{blue}{GROW OR} DISAPPEAR \\
& \tiny Male & \tiny \textcolor{red}{THE DEBUT} COMPANIES TO \textcolor{red}{GO ON} DISAPPEAR \\
& \tiny Male$\rightarrow$Female & \tiny \textcolor{blue}{THEY EXPECT} COMPANIES TO \textcolor{blue}{GROW OR} DISAPPEAR \\
\cmidrule{2-3}
& \tiny True & \tiny MR \textcolor{blue}{POLO} ALSO \textcolor{blue}{OWNS} THE FASHION COMPANY   \\
& \tiny Male & \tiny MR \textcolor{red}{PAYING} ALSO \textcolor{red}{LONG} THE FASHION COMPANY \\
& \tiny Male$\rightarrow$Female & \tiny MR \textcolor{blue}{POLO} ALSO \textcolor{blue}{OWNS} THE FASHION COMPANY \\
\bottomrule
\end{tabular}
\vspace{-1.5em}
\end{table}
\renewcommand{\arraystretch}{1}
% Most of the variations across different spectrogram domains can be represented as per frequency band  variations. 

To accommodate generative adversarial network for training on non-parallel spectrogram domains, the generator should be back-propagated with multiple gradient signals (from different discriminators), that each represents the variations between source and target domains at a specific frequency band. 
% To accommodate CycleGAN generators with such gradient signals, 
To achieve this goal, we propose to use multiple and independent discriminators for each domain, similar to generative multi adversarial network (GMAN)~\cite{Durugkar2016GenerativeMN}. 
We show that the proposed Multi-Discriminator CycleGAN, without pretraining the discriminators, outperforms CycleGAN~\cite{CycleGAN2017} with pretrained discriminator, for spectrogram adaptation. 
% YZ: I think the "It is shown" should be "We show"
Furthermore, we show that the multi discriminator architecture can overcome the checkerboard artifacts problem caused by deconvolution layer in generator~\cite{olah2016deconv}, and generates natural clean audio. To evaluate the performance of the proposed model, gender-based domains are selected as domain adaptations. 
% YZ: same as above, the "It is shown" should be "We show"
% Recently, generative models have been widely used for speech domain adaptation. Autoregressive generative models have been used for raw audio and text-to-speech (TTS) synthesis. Wavenet~\cite{Oord2016WaveNetAG,Paine2016FastWG,Oord2017ParallelWF} and SampleRNN~\cite{Mehri2016SampleRNNAU} have been developed to directly synthesize audio samples using recurrent and convolutional networks. Shen et al.~\cite{Shen2017NaturalTS} proposed to conditioning Wavenet on mel-spectrogram, to directly synthesize speech from text.

% YZ: these are all TTS models, which is not really relevant
\vspace{-0.5em}
\subsection{Related Work}
\vspace{-0.5em}
Generative Adversarial Network (GAN)~\cite{goodfellow14_gan} is a family of non-parametric density estimation models which learn to model the data generating distribution using adversarial training. Conditional GANs (CGAN)~\cite{MirzaO14} was first proposed for supervised (parallel) domain adaptation, where the goal is to convert source distribution to match the target. %using adversarial training. 
CGAN has been used in various data domains, especially image domains, both for parallel~\cite{pix2pix2017,pix2pixHD2017} and non-parallel domain adaptation~\cite{DiscoGAN2017,DualGAN2017,CycleGAN2017}. 

Recently, CGAN is used for speech enhancement on parallel datasets~\cite{MirzaO14,pix2pix2017}. Speech denoising is achieved by conditioning the generator on noisy speech to learn the de-noised version~\cite{Pascual2017SEGANSE,Michelsanti2017ConditionalGA}. Donahue et al.~\cite{Donahue2018SynthesizingAW} proposed a GAN model on audio (WaveGAN) and spectrogram (SpecGAN), which is actually trained CGAN on parallel domains. %Recently, VC model based on GAN's has been proposed~\cite{kaneko17-cycgan,hsu17-vae,hsu17-wgan}. 
Kaneko et al.~\cite{kaneko17-cycgan} proposed a cycle-consistent adversarial network (CycleGAN)~\cite{CycleGAN2017} with gated convolutional neural network (CNN) as the generator part, where the model is trained on Mel-cepstral coefficients (MCEPs) features. Hsu et al.~\cite{hsu17-wgan} proposed a combination of variational inference network, using variational autoencoder (VAE)~\cite{VAE2013}, and adversarial network, using Wasserstein GAN (WGAN)~\cite{WassersteinGAN2017}. In~\cite{hsu17-wgan}, the goal is to disentangle the linguistic from nuisance latent variables via VAE using spectra (SP for short), aperiodicity (AP), and pitch contours (F0) features, followed by adversarial training to learn the target distribution from the inferred linguistic latent distribution. A recurrent VAE is also proposed~\cite{hsu17-vae,Hsu2017UnsupervisedLO} to capture the temporal relationships in the disentangled representation of sequence data, using Mel-scale filter bank (FBank).

% Moreover, the multiple gradients should be computed independently to represent the frequency band variations through time domains. 

% to compute and back-propagate the intended gradients
Contributions of the proposed generative model are, \textbf{(1)} It is a robust GAN model developed for non-parallel unsupervised domains, compared to parallel-based SpecGAN and WaveGAN~\cite{Donahue2018SynthesizingAW}, \textbf{(2)} The choice of multiple discriminator is adjustable to the spectro-temporal structure of the intended domains, compared to domain-specific model design of~\cite{kaneko17-cycgan}, \textbf{(3)} Proposed GAN model training is robust and invariant to the choice of adversarial objective, i.\,e. binary cross-entropy or least square (LS-GAN~\cite{LSGAN2017}), while the CycleGAN in~\cite{kaneko17-cycgan} is only stable using least square 
% YZ: do you mean "while the CycleGAN" ? it is very confusing, as in both places you mentioned proposed CycleGAN
loss, with additional using of identity mapping loss in generator, \textbf{(4)} Source and target domains in~\cite{kaneko17-cycgan} are sampled from same speakers, both including male and female, only uttering different sentences, while our approach is more natural as source and targets distribution is strongly diverged due to different speaker, gender, and uttered sentences. \textbf{(5)} Compared to FHVAE~\cite{Hsu2017UnsupervisedLO}, our models improves ASR performance on TIMIT female set by $2.067\%$ PER (Table~\ref{tab:timit-train-male}), when only trained on male. 

% \begin{description}
% \item[(1)] It is a robust GAN model developed for non-parallel unsupervised domains, compared to parallel-based SpecGAN and WaveGAN~\cite{Donahue2018SynthesizingAW} 
% \item[(2)] The choice of multiple discriminator is adjustable to the spectro-temporal structure of the intended domains whereas~\cite{kaneko17-cycgan} model design is domain specific 
% \item[(3)] Training of the proposed GAN model is robust and invariant to the choice of adversarial objective, i.\,e. negative log-likelihood or least square (LS-GAN~\cite{LSGAN2017}), while the proposed CycleGAN in~\cite{kaneko17-cycgan} is only stable using least square loss, with using identity mapping loss in generator 
% \item[(4)] Source and target domains in~\cite{kaneko17-cycgan} are sampled from same speakers, both including male and female, only uttering different sentences, while our approach is more natural as source and targets distribution is strongly diverged due to different speaker, gender, and uttered sentences. 
% \end{description}

% This paper is organized as follows. In Section~\ref{sec:proposed-model} the proposed model is described, by first reviewing the GAN and CGANs (Section~\ref{sebsec:domain-adaptation}), followed by proposing Multi-Discriminator CycleGAN in Section~\ref{subsec:md-cyclegan}. Section~\ref{sec:experiment} describes the experiments with qualitative (Section~\ref{subsec:qualitative}) and quantitative evaluations(Section~\ref{subsec:quantitative}) of generated samples. Finally, Section~\ref{sec:conclusion} explain the conclusion and future directions. 

\vspace{-1em}
\section{Proposed Model}
\label{sec:proposed-model}
\vspace{-0.5em}
In this section, the proposed model is explained. We first describe the generative model based on adversarial network. Generative Model based on adversarial training (GAN) has been proposed by Goodfellow et al.~\cite{goodfellow14_gan} to model the data generating distribution. Training GAN is based on minimizing Jensen-Shannon divergence between data generating distribution $p_{data}(x)$ and model data distribution $p_z(z)$. Learning is through minimization of the adversarial loss between generator network $G(z)$, which learns a mapping function $G: Z\rightarrow X$, and discriminator network $D(x)$. The generator is learning to model the data distribution $p_{data}(x)$ by generating indistinguishable samples $\hat{x}=G(z)$ from $x$, using a source noise signal $z$ to minimize (\ref{eq:gan}),  whereas discriminator is learning to discriminate between real data $x$ and generated $\hat{x}$ by maximizing the adversarial loss,
\begin{equation}
\begin{split}
\label{eq:gan}
\mathcal{L}_{GAN}\left(G, D\right) = & \mathbb{E}_{x\sim p_{data}(x)}\left[\log D(x)\right] +  \\
& \mathbb{E}_{z\sim p_{z}(z)}\left[\log \left(1-D\left(G(z)\right)\right)\right] 
\end{split}
\end{equation}
\vspace{-2em}
\subsection{Domain Adaptation via GAN}
\label{sebsec:domain-adaptation}
\vspace{-0.5em}
For domain adaptation between parallel domains $X$ and $Y$, Conditional GAN (CGAN)\cite{MirzaO14,pix2pix2017} is proposed, using a generator that directly learns the mapping function $G: X\rightarrow Y$, by minimizing parallel conditional adversarial loss $\mathcal{L}_{P-CGAN}$, 
\vspace{-1em}
\begin{equation}
\begin{split}
\label{eq:cgan-parallel}
&\mathcal{L}_{P-CGAN}\left(G, D\right) = \mathbb{E}_{(x,y)\sim p_{data}(x,y)}\left[\log D(x,y)\right] + \\
& \mathbb{E}_{x\sim p_{data}(x), z\sim p_{z}(z)}\left[\log \left(1-D\left(x,G(z,x)\right)\right)\right] 
\end{split}
\end{equation}
where $D$ is discriminating between pair of real parallel data $(x,y)$ and generated pair $(x,G(z,x))$.
To apply CGAN for adaptation between non-parallel domains $X$ and $Y$, a conditional GAN using cycle consistent adversarial loss (CycleGAN) has been proposed~\cite{CycleGAN2017,DiscoGAN2017,DualGAN2017}. In CycleGAN~\cite{CycleGAN2017}, there are two conditional generators, i.\,e., $G_{X}:X\rightarrow Y$ and $G_{Y}:Y\rightarrow X$, each trained in adversarial setting with $D_{Y}$ and $D_{X}$, respectively. In other words, there are two pairs of Non-parallel conditional adversarial loss $\mathcal{L}_{NP-CGAN}(G_{X}, D_{Y})$ and $\mathcal{L}_{NP-CGAN}(G_{Y}, D_{X})$, where, 
\begin{equation}
\begin{split}
\label{eq:cgan-unparallel}
&\mathcal{L}_{NP-CGAN}\left(G_{X}, D_{Y}\right) = \mathbb{E}_{(y)\sim p_{Y}(y)}\left[\log D_{Y}(y)\right] + \\
& \mathbb{E}_{x\sim p_{X}(x), z\sim p_{z}(z)}\left[\log \left(1-D_{Y}\left(G_{X}(z,x)\right)\right)\right] 
\end{split}
\end{equation}
In non-parallel situation, the goal is to find the correct pseudo pair $(x,y)$ across $X$ and $Y$ domains in an unsupervised way. To ensure that $G_{X}$ and $G_{Y}$ will learn such mapping function, CycleGAN\cite{CycleGAN2017} proposed to minimize a cycle consistency loss using $\ell_{1}$ norm,
\begin{equation}
\begin{split}
\label{eq:cycle}
\mathcal{L}_{cycle} = &\mathbb{E}_{x\sim p_{X}(x)}\left[\parallel G_{Y}(G_{X}(x))-x\parallel_{1} \right]+\\
&\mathbb{E}_{y\sim p_{Y}(y)}\left[\parallel G_{X}(G_{Y}(y))-y\parallel_{1} \right]
\end{split}
\end{equation}
Therefore, CycleGAN\cite{CycleGAN2017} learns unsupervised mapping functions between $X$ and $Y$ domains by combining~(\ref{eq:cgan-unparallel}) and~(\ref{eq:cycle}), to maximize the adversarial loss $\mathcal{L}_{CycleGAN}$, where, 
\begin{equation}
\begin{split}
\label{eq:cycgan}
\mathcal{L}_{CycleGAN} = & \mathcal{L}_{NP-CGAN}(G_{X}, D_{Y}) + \\
					     & \mathcal{L}_{NP-CGAN}(G_{Y}, D_{X}) + \\
                         & -\lambda\mathcal{L}_{cycle}(G_{X}, G_{Y}) \\
\end{split}
\end{equation}

\vspace{-1em}
\subsection{Multi-Discriminator CycleGAN (MD-CycleGAN)}
\label{subsec:md-cyclegan}
\vspace{-0.5em}
In this section, we propose a multiple discriminator generative model based on cycle consistency  loss~(\ref{eq:cycgan}). The model is based on generative multi adversarial network (GMAN)~\cite{Durugkar2016GenerativeMN}. In this paper, $X$ and $Y$ represents spectrogram feature datasets of different speech domains. Spectrogram feature represents the frequency variation of audio data through time dimension. In order to allow CycleGAN to learn the mapping function of spectrogram between different speech domains, the generators $\left\lbrace G_{X},G_{Y}\right\rbrace$ should be able to learn the variations in each frequency band for each aligned time window, across domains.
% $X$ and $Y$ domains. 

In order to learn the frequency-dependent mapping functions $\left\lbrace G_{X}, G_{Y}\right\rbrace$ that catch the variation per each frequency bands, we define multiple frequency-dependent discriminators $\left\lbrace D_{X}^{f_{j\in n}}, D_{Y}^{f_{i\in m}}\right\rbrace$, where $f_{j\in n}$ represents the $i^{th}$ frequency band of domain $X$ with $n$ frequency bands, and $f_{i\in m}$ represents $j$-th frequency band od domain $Y$, respectively. The frequency band definition in each domain can share a portion of frequency spectrum, or be exclusive, based on the domain spectrogram distribution. We are also using the non-saturating version of GAN\cite{goodfellow14_gan}, NS-GAN, where the generator $G$ is learned through maximizing the probability of predicting generated samples $\hat x$ as drawn from data generating distribution $p_{data}(x)$. Accordingly, the adversarial loss for each pair of generator and discriminator $\left\lbrace \left(G_{X}, D^{f_{i\in m}}_{Y}\right), \left(G_{Y}, D^{f_{j\in n}}_{X}\right)\right\rbrace$ in~(\ref{eq:cgan-unparallel}) and~(\ref{eq:cycgan}) is
\begin{equation}
\begin{split}
\label{eq:cgan-multiD}
&\mathcal{L}_{MD-CGAN}\left(G_{X}, D^{f_{i\in m}}_{Y}\right) = \mathbb{E}_{(y)\sim p_{Y}(y)}\left[\sum_{i=1}^{m}\log D^{f_{i}}_{Y}(y)\right] \\
& + \mathbb{E}_{x\sim p_{X}(x), z\sim p_{z}(z)}\left[\sum_{i=0}^{m}\log \left(D^{f_{i}}_{Y}\left(G_{X}(z,x)\right)\right)\right] 
\end{split}
\end{equation}

The Multi-Discriminator CycleGAN (MD-CycleGAN) is training by maximizing $\mathcal{L}_{MD-CycleGAN}$, where,
\begin{equation}
\begin{split}
\label{eq:md-cycgan}
\mathcal{L}_{MD-CycleGAN} = & \mathcal{L}_{MD-CGAN}(G_{X}, D^{f_{i\in m}}_{Y}) + \\
					     & \mathcal{L}_{MD-CGAN}(G_{Y}, D^{f_{j\in n}}_{X}) + \\
                         & -\lambda\mathcal{L}_{cycle}(G_{X}, G_{Y}) \\
\end{split}
\end{equation}
A natural extension to the proposed MD-CycleGAN is to use multiple frequency-dependent generators~\cite{Ghosh2017MultiAgentDG} jointly with discriminators as well. This can follow in two configurations. In one-one setting, each generator is trained on a specific frequency band with the corresponding discriminator, i.\,e., set of $\left\lbrace \left(G_{X}^{f_{i}}, D_{Y}^{f_{i}}\right): i\in m\right\rbrace$. Additionally, in one-many setting, each frequency-dependent generator is trained with all frequency-dependent discriminators, i.\,e., set of $\left\lbrace \left(G_{X}^{f_{j}}, D_{Y}^{f_{i\in m}}\right): j\in n\right\rbrace$ trained in adversarial setting.

% \vspace{-0.5em}
\section{Experiment}
\label{sec:experiment}
% \vspace{-0.5em}
We used TIMIT ~\cite{timit} and Wall Street Journal (WSJ) corporas to evaluate the performance of proposed model on domain adaptation. TIMIT dataset contains broadband
$16$kHz recordings of phonetically-balanced read speech of $6300$ utterances ($5.4$ hours). Male/Female ratio of speakers across train/validation/test sets are approximately $70$\% to $30$\%. WSJ contains $\approx80$ hours of standard \emph{si284}/\emph{dev93}/\emph{eval92} for train/validation/test sets, with equally distributed genders.

% TIMIT $10$ sentence for each of $630$ speakers

% WSJ dataset contains $38252$ utterances ($\approx80$ hours) of standard \emph{si284} set for training, \emph{dev93} for validation and \emph{eval92} for test evaluation, containing $100/40/50$ sentences per speaker for train, validation, and test sets, respectively.  Male/Female is equally distributed in WSJ dataset. containing $100/40/50$ sentences per speaker for train, validation, and test sets, respectively. \emph{LDC93S6B} and \emph{LDC94S13B},

The spectrogram representation of audio is used for training the CycleGAN and ASR models, which is computed with a $20$ms window and $10$ms step size. Each spectrogram is normalized to have zero mean and unit variance. 
% No further preprocessing is done after normalization.
To implement MD-CycleGAN, three non-overlapping frequency bands are defined, i.\,e. $m=n=3$ with $[53, 53, 55]$ bandwidth, for male and female domains. 
% We have also tried up to 6 frequency band and did not notices a significant improvement on audio quality. 
We denote the size of the convolution layer by the tuple $(\text{C, F, T, SF, ST})$, where C, F, T, SF, and ST denote number of channels, filter size in frequency dimension, filter size in time dimension, stride in frequency dimension and stride in time dimension respectively. CycleGAN model architecture is based on~\cite{CycleGAN2017} with some modifications. The generator is based on U-net~\cite{UNet2015} architecture with $4$ convolutional layers of sizes (8,3,3,1,1), (16,3,3,1,1), (32,3,3,2,2), (64,3,3,2,2) with corresponding deconvolution layers. We noticed that the discriminator in~\cite{CycleGAN2017} outputs a vector with dimension equal to the channel size of final convolution layer, instead of outputting a scalar~\cite{goodfellow14_gan}. It was observed that this causes instability in a balanced training between generator and discriminator. We modified this by adding a fully connected layer as final layer, to match the discriminator in~\cite{goodfellow14_gan}.
% output a scalar of $0$ or $1$ as fake or true sample. 
Discriminator has $4$ convolution layers of sizes (8,4,4,2,2), (16,4,4,2,2), (32,4,4,2,2), (64,4,4,2,2), as default kernel and stride sizes in~\cite{CycleGAN2017}. We used Griffin-lim algorithm~\cite{Griffin1983SignalEF} for audio reconstruction, to assess its quality. ASR model is based on~\cite{zhou2017improving}, trained with maximum likelihood, and no policy gradient. The model has one convolutional layer of size (32,41,11,2,2), and five residual convolution blocks of size (32,7,3,1,1), (32,5,3,1,1), (32,3,3,1,1), (64,3,3,2,1), (64,3,3,1,1) respectively. Following the convolutional layers are $4$ layers of bidirectional GRU RNNs with $1024$ hidden units per direction per layer, one fully-connected hidden layer of size $1024$ and final output layer. 

\vspace{-0.5em}
\subsection{Quantitative Evaluation}
\label{subsec:quantitative}
% \vspace{-0.5em}
In this section, ASR model is employed to evaluate the performance of proposed model, where domains are different genders. 
% The goal of domain adaptation is to improve the performance of ASR model by adapting the distribution of target domain and source, where we define domains differentiated by gender type. 
First, gender generators $\left\lbrace G_{M\rightarrow F}, G_{F\rightarrow M}\right\rbrace$ \footnote{{\tiny $G_{M\rightarrow F}: Male\rightarrow Female$, $G_{F\rightarrow M}: Female\rightarrow Male$}} are trained on gender-separated train set. These generators are then evaluated for train$\rightarrow$test and test$\rightarrow$train adaptation using ASR model. In former, ASR model is retrained on the adapted train set, while in latter, a more applicable case, ASR model is fixed and evaluated on the new adapted test sets.

\begin{table}[tbp!]
\caption{TIMIT, Train set Female$\rightarrow$Male domain adaptation. \emph{Note:} Female\&$\rightarrow$Male means Female+Female$\rightarrow$Male}
\label{tab:timit-train-female}
\vspace{-1em}
\centering
\begin{tabular}{llcc}
\toprule
 &  &  \multicolumn{2}{c}{Male (PER)} \\
Model & Train & Val & Test\\
\toprule
 & \small Female & 40.704 & 42.788\\
 \midrule
\multirow{2}{*}{{\small One-D CycleGAN}} & \small Female$\rightarrow$Male & 40.095 & 42.379\\
& \small Female\&$\rightarrow$Male & 39.200 & 42.211\\
\midrule
\multirow{2}{*}{\small Three-D CycleGAN} & \small Female$\rightarrow$Male & \textbf{29.838} & 33.463\\
& \small Female\&$\rightarrow$Male & 30.009 & \textbf{33.273}\\
\midrule
& \small Male (baseline) & 20.061 & 22.516\\
\bottomrule
\end{tabular}
\vspace{-0.5em}
\end{table}

\begin{table}[tbp!]
\caption{TIMIT, Train set Male$\rightarrow$Female domain adaptation. \emph{Note:} Male\&$\rightarrow$Female means Male+Male$\rightarrow$Female}
\label{tab:timit-train-male}
\vspace{-1em}
\centering
\begin{tabular}{llcc}
\toprule
 &  &  \multicolumn{2}{c}{Female (PER)} \\
Model & Train & Val & Test\\
\toprule
 & \small Male & 35.702 & 30.688\\
 \midrule
\multirow{2}{*}{{\small One-D CycleGAN}} & \small Male$\rightarrow$Female & 32.943 & 30.069\\
& \small Male\&$\rightarrow$Female & 31.289 & 29.038\\
\midrule
\multirow{2}{*}{\small Three-D CycleGAN} & \small Male$\rightarrow$Female & 28.80 & 25.448 \\
& \small Male\&$\rightarrow$Female & \textbf{25.982} & \textbf{24.133} \\
\midrule
FHVAE~\cite{Hsu2017UnsupervisedLO} & Male + $\mathit{\mathbf{z}}_{1}$ & & 26.20 \\
\midrule
& \small Female (baseline) & 24.51 & 23.215\\
\bottomrule
\end{tabular}
\vspace{-0.5em}
\end{table}

\begin{table}[tbp!]
\caption{WSJ, Train set Female$\leftrightarrow$Male domain adaptation, using Three-D CycleGAN trained on TIMIT train set.}
% The Male and Female generators are not retrained on WSJ dataset.}
  \label{tab:wsj-train-adaptation}
  \vspace{-1em}
\centering
\begin{tabular}{lcccc}
\toprule
 &  \multicolumn{4}{c}{Test -\emph{eval92}} \\
 & \multicolumn{2}{c}{Male} & \multicolumn{2}{c}{Female}\\
 Train & CER & WER & CER & WER\\

\toprule
\small Female (baseline) & 14.31 & 27.66  & 2.80 & 6.71 \\
\small Female\&$\rightarrow$Male & \textbf{5.20} & \textbf{12.39} &  \\
% &  &  & \\
\midrule
\small Male (baseline) & 3.19 & 8.16 & 7.57 & 16.38 \\
 \small Male\&$\rightarrow$Female & & & \textbf{4.22} & \textbf{9.46} \\
% &   &  &  \\
\bottomrule
\end{tabular}
\vspace{-1em}
\end{table}

\begin{figure*}[h!]
\vspace{-3em}
\centering
 \subfloat[Female]{
   \includegraphics[width=0.35\textwidth]{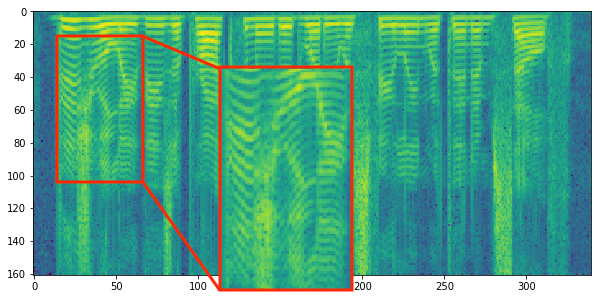}
   \label{fig:female-real}
 }
 \subfloat[Male]{
   \includegraphics[width=0.465\textwidth]{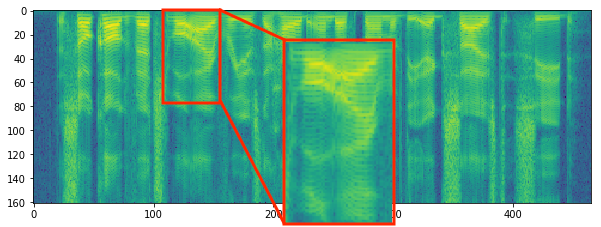}
   \vspace{-1em}
   \label{fig:male-real}
 }
%  \qquad
 \vspace{-1.5em}

 \subfloat[Female $\rightarrow$Male, One-D CycleGAN (Pretrained)]{
   \includegraphics[width=0.35\textwidth]{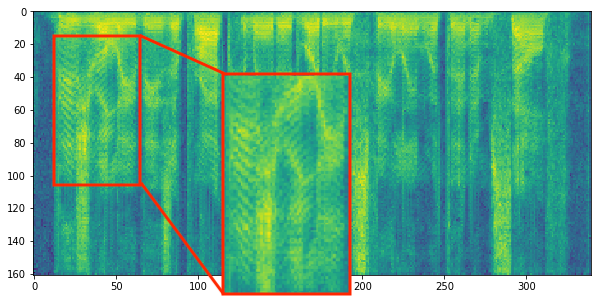}
   \label{fig:female-male-oneD}
 }
 \subfloat[Male $\rightarrow$ Female, One-D CycleGAN (Pretrained)]{
   \includegraphics[width=0.465\textwidth]{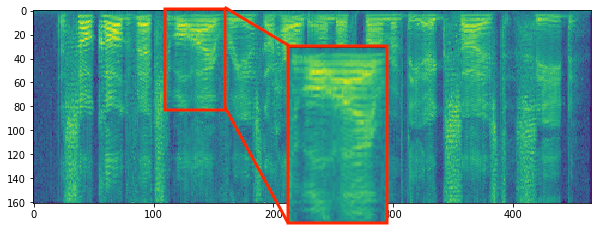}
   \label{fig:male-female-oneD}
 }
%  \qquad
 \vspace{-1em}

 \subfloat[Female $\rightarrow$ Male, Three-D CycleGAN]{
   \includegraphics[width=0.35\textwidth]{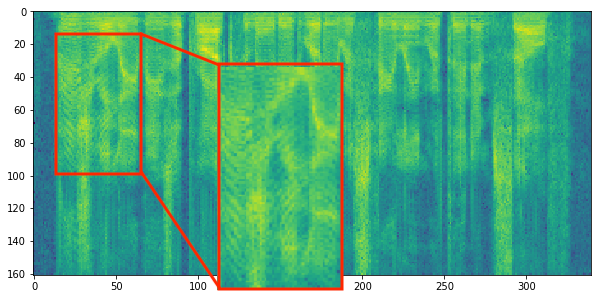}
   \label{fig:female-male-threeD}
 }
 \subfloat[Male $\rightarrow$ Female, Three-D CycleGAN]{
   \includegraphics[width=0.465\textwidth]{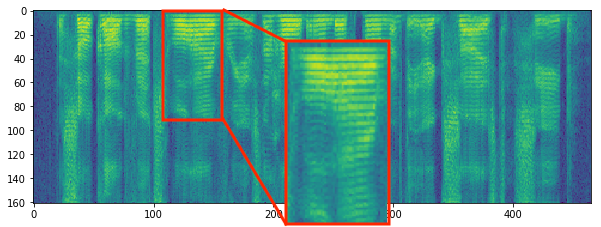}
   \label{fig:male-female-threeD}
 }
%  \vspace{-1em}
 \caption{Spectrogram conversion for (a,c,e) female$\rightarrow$male, and (b,d,f) male$\rightarrow$female, using One-D CycleGAN and Three-D CycleGAN on TIMIT test set. \textbf{Note:} The One-D CycleGAN generator converges only by pretraining the discriminator first, unless the generator will learn identity mapping function. However, the Three-D CycleGAN results are achieved without pretraining.}
 \label{fig:spectrogram-compare}
\vspace{-1.5em}
\end{figure*}

\vspace{-0.5em}
\subsubsection{Train$\rightarrow$Test Adaptation}
\label{subsubsec:train_adapt}
\vspace{-0.5em}
Results on adapting TIMIT train set are shown in Table~\ref{tab:timit-train-female} and~\ref{tab:timit-train-male}. As ablation study to CycleGAN-VC~\cite{kaneko17-cycgan}, performance is significantly improved with three discriminator compared to single one. Compared to FHVAE~\cite{Hsu2017UnsupervisedLO}, phoneme error rate is improved by $2.067\%$ in Table~\ref{tab:timit-train-male}.  
To evaluate the generalization of the generators, we used them on WSJ dataset without retraining. As shown in Table~\ref{tab:wsj-train-adaptation}, ASR performance is significantly improved by reducing the gap to the corresponding male and female baselines. For a fair comparison, ASR performance trained on WSJ train set is $5.55\%$ WER. It is worth mentioning that relatively lower performance on TIMIT is due to smaller size of dataset.

\vspace{-0.5em}
\subsubsection{Test$\rightarrow$Train Adaptation}
\label{subsubsec:test_adapt}
\vspace{-0.5em}
Test set adaptation of TIMIT and WSJ are shown in Table~\ref{tab:timit-test-male-female} and~\ref{tab:wsj-test-male-female}. It is clear that using the proposed model, ASR performance is significantly improved by adapting test$\rightarrow$train 
% distribution
, compared to original CycleGAN. Qualitative assessment of ASR predictions are shown in Tables~\ref{tab:wsj-asr} and Appendix~\ref{appendix-asr}.
% ~\ref{tab:wsj-asr-male-1},~\ref{tab:wsj-asr-female-1},~\ref{tab:wsj-asr-male-2},~\ref{tab:wsj-asr-female-2}.

\vspace{-0.5em}
\begin{table}[tbh!]
\caption{TIMIT, Test set Male$\leftrightarrow$Female domain adaptation}
  \label{tab:timit-test-male-female}
  \vspace{-1em}
\centering
\begin{tabular}{llcc}
\toprule
 &  &  \multicolumn{2}{c}{Train} \\
Test (PER) & Model & Male & Female\\
\toprule
\small Male (baseline) & -- & 22.516  & 42.788 \\
%  \midrule
\multirow{2}{*}{{\small Male$\rightarrow$Female}} & \small \small One-D CycleGAN &  & 43.427 \\
& \small Three-D CycleGAN &  & \textbf{37.000} \\
\midrule
\small Female (baseline) & -- & 32.085 & 23.215\\
\multirow{2}{*}{\small Female$\rightarrow$Male} & \small One-D CycleGAN & 32.606 & \\
& \small Three-D CycleGAN & \textbf{25.758} &  \\

\bottomrule
% \bottomrule
\end{tabular}
\vspace{-1em}
\end{table}

\begin{table}[tbh!]
\caption{WSJ, Test set Male$\leftrightarrow$Female domain adaptation}
\vspace{-1em}
  \label{tab:wsj-test-male-female}
\centering
\begin{tabular}{lcc}
\toprule
 & \multicolumn{2}{c}{Train} \\
Test (CER / WER) & Male & Female\\
\toprule
\small Male (baseline) & 3.19 / 8.16  & 14.31 / 27.66 \\
%  \midrule
\small Male$\rightarrow$Female & & \textbf{6.82} / \textbf{15.68} \\
\small Female (baseline) & 7.57 / 16.38 & 2.80 / 6.71\\
\small Female$\rightarrow$Male & \textbf{5.93} / \textbf{13.18} & \\
\bottomrule
\end{tabular}
\vspace{-1.5em}
\end{table}

% \vspace{-0.5em}
\subsection{Qualitative Evaluation}
\label{subsec:qualitative}
% \vspace{-0.5em}
In this section, the quality of generated spectrogram for male$\leftrightarrow$female adaptation is assessed. The characteristic difference between male and female spectrograms is the variation rate of frequency for a fixed time window. As shown in Figure~\ref{fig:spectrogram-compare}, top row depicts the original spectrograms, where male is characterized by smooth frequency variation, opposed to peaky and high-rate variations of female. Well-trained generators should catch these inter-domain variations. As ablation study, we are also showing the generated spectrogram by CycleGAN~\cite{CycleGAN2017} (One-D CycleGAN), in middle row, comparing with Three-D CycleGAN in bottom row. One-D CycleGAN learns to convert the spectrogram only using a pretrained discriminator. It is noticeable that the converted spectrogram in One-D CycleGAN fails to match the target domain characteristics, at some frequency bands, and simply copied the source spectrogram. However, with no pretraining of Three-D CycleGAN, it learns a better mapping function, by either suitably smoothing the spectrogram (female$\rightarrow$male), or generating peaky variations (male$\rightarrow$female). The checkerboard artifacts~\cite{olah2016deconv} is a common problem in deconvolution-based generators. This problem is visible in One-D CycleGAN, with discontinuous artifacts through time and frequency dimensions, which results in a noisy and unnatural-sounding audio. This problem is mitigated in Three-D CycleGAN, by learning the target domain characteristics using multiple independent discriminators. 
% To further investigate the audio, the reconstructed audio's of TIMIT and WSJ for both models are provided here[link].

\vspace{-0.5em}
\section{Conclusion and Future Directions}
\label{sec:conclusion}
\vspace{-0.5em}
In this paper, a new cyclic consistent generative adversarial network based on multiple discriminators is proposed (MD-CycleGAN) for unsupervised non-parallel speech domain adaptation. Based on the frequency variation of spectrogram between domains, the multiple discriminators enabled MD-CycleGAN to learn an appropriate mapping functions that catch the frequency variations between domains. The performance of MD-CycleGAN is measured by ASR prediction, when train and test set are sampled from different genders. 
% on source$\leftrightarrow$target, based on gender difference. 
% Evaluated on gender-based domains, 
It was shown that MD-CycleGAN can improve the ASR performance on unseen domains. As future extension, this model will be evaluated on datasets adaptation, e.\,g. TIMIT$\leftrightarrow$WSJ, and accent, e.\,g. American$\leftrightarrow$Indian adaptations. 

\bibliographystyle{IEEEtran}

\bibliography{mybib}

\begin{appendices}
% \appendix

\section{ASR Prediction Assessment}
\label{appendix-asr}
% \vspace{-5em}
In this section, qualitative assessment of ASR predictions are shown, when test set is sampled from the opposite gender, and when MD-CycleGAN model is used to adapt the test set distribution to the train set. Tables~\ref{tab:wsj-asr-male-1},~\ref{tab:wsj-asr-male-2} show the selected ASR predictions when adapting female$\rightarrow$male, whereas Tables~\ref{tab:wsj-asr-female-1},~\ref{tab:wsj-asr-female-2} show adaptation for male$\rightarrow$female. Note that MD-CycleGAN is trained on TIMIT, and used for adaptation on WSJ dataset.

\vspace{-0.5em}
\begin{table}[h!]
\caption{ASR prediction mismatch when train on male and evaluated on female (WSJ dataset).}
  \label{tab:wsj-asr-male-1}
  \vspace{-1em}
\centering
\begin{tabular}{p{0.65cm}p{0.85cm}p{5.5cm}}
\toprule
 &  & \scriptsize Train on Male  \\
\midrule
\multirow{6}{1cm}{\scriptsize Test on Female} & \tiny True & \tiny \textcolor{blue}{ASSOCIATED INNS KNOWN AS AIRCOA} IS THE GENERAL PARTNER OF \textcolor{blue}{AIRCOA HOTEL PARTNERS AND HAS} A \ldots 
% ONE PERCENT GENERAL PARTNER INTEREST IN THE 
PARTNERSHIP  \\
& \tiny Female & \tiny \textcolor{red}{AFFECTED ENDS NONE IS AIR COLA} IS THE GENERAL PARTNER OF \textcolor{red}{ ALOHA TELL PRINTERS} \textcolor{blue}{AND HAS A} \ldots 
% ONE PERCENT GENERAL PARTNER INTEREST IN THE 
PARTNERSHIP \\
& \tiny Female$\rightarrow$Male & \tiny \textcolor{blue}{ASSOCIATED INNS} \textcolor{red}{ NONE HAS} \textcolor{blue}{ AIRCOA} IS THE GENERAL PARTNER OF \textcolor{red}{THE ARCO} \textcolor{blue}{HOTEL PARTNERS AND} \textcolor{red}{AS} A 
% ONE PERCENT GENERAL PARTNER INTEREST IN THE 
PARTNERSHIP \\

\cmidrule{2-3}
& \tiny True & \tiny \textcolor{blue}{ALTHOUGH THOSE GAINS ERODED} DURING \ldots 
% THE AFTERNOON STOCK PRICES STAYED WITHIN A NARROW RANGE 
UNTIL THE LAST \textcolor{blue}{HALF HOUR} OF TRADING \\
& \tiny Female & \tiny \textcolor{red}{ ALL THE THIS} \textcolor{blue}{GAINS} \textcolor{red}{A ROLE} DURING \ldots 
% THE AFTERNOON STOCK PRICES STAYED WITHIN A NARROW RANGE 
UNTIL THE LAST \textcolor{red}{THATCHER} OF TRADING \\
& \tiny Female$\rightarrow$Male & \tiny \textcolor{blue}{ALTHOUGH THOSE} \textcolor{blue}{GAINS} \textcolor{red}{A ROLE} DURING \ldots 
% THE AFTERNOON STOCK PRICES STAYED WITHIN A NARROW RANGE 
UNTIL THE LAST \textcolor{red}{TOUGH} \textcolor{blue}{HOUR} OF TRADING \\

\cmidrule{2-3}
& \tiny True & \tiny \textcolor{blue}{LA GUARDIA} HAS ONLY FIFTY SEVEN GATES \textcolor{blue}{BUT} AT PEAK HOURS DOZENS OF MORE PLANES \textcolor{blue}{MAY BE ON THE} GROUND \\
& \tiny Female & \tiny \textcolor{red}{THE GUIDE A} HAS ONLY FIFTY SEVEN GATES \textcolor{red}{THAT} AT PEAK HOURS DOZENS OF MORE PLANES \textcolor{red}{NAVY} \textcolor{blue}{ON} \textcolor{red}{A} GROUND \\
& \tiny Female$\rightarrow$Male & \tiny \textcolor{red}{THE GUARD A} HAS ONLY FIFTY SEVEN GATES \textcolor{blue}{BUT} AT PEAK HOURS DOZENS OF MORE PLANES \textcolor{blue}{MAY BE ON} \textcolor{red}{A} GROUND \\

\cmidrule{2-3}
& \tiny True & \tiny HE \textcolor{blue}{SAID THE SALES HAD} HAD A MAJOR PSYCHOLOGICAL \textcolor{blue}{IMPACT} ON IRAN AND A NEGATIVE MILITARY \textcolor{blue}{IMPACT ON IRAQ} \\
& \tiny Female & \tiny HE \textcolor{red}{FED THE FAILS AND} HAD A MAJOR PSYCHOLOGICAL \textcolor{red}{INTACT} ON IRAN AND A NEGATIVE MILITARY \textcolor{red}{INTACT UNDER} \\
& \tiny Female$\rightarrow$Male & \tiny HE \textcolor{blue}{SAID THE SALES HAD} HAD A MAJOR PSYCHOLOGICAL \textcolor{blue}{IMPACT} ON IRAN AND A NEGATIVE MILITARY \textcolor{blue}{IMPACT ON IRAQ} \\

\cmidrule{2-3}
& \tiny True & \tiny HE SAYS IT CAN GET \textcolor{blue}{CORNY TO SAY THAT} MUSIC IS A UNIVERSAL LANGUAGE \textcolor{blue}{BUT IT} REALLY IS \\
& \tiny Female & \tiny HE SAYS IT CAN GET \textcolor{red}{CORN TO SANTA} MUSIC IS A UNIVERSAL LANGUAGE \textcolor{red}{THAT} \textcolor{blue}{IT} REALLY IS \\
& \tiny Female$\rightarrow$Male & \tiny HE SAYS IT CAN GET \textcolor{red}{CORNER} \textcolor{blue}{TO SAY} \textcolor{red}{THE} MUSIC IS A UNIVERSAL LANGUAGE \textcolor{blue}{BUT} \textcolor{red}{A} REALLY IS \\

\cmidrule{2-3}
& \tiny True & \tiny \textcolor{blue}{GLASNOST} \ldots 
% HAS ALSO 
\textcolor{blue}{BEEN} GOOD TO LAWRENCE \textcolor{blue}{LEIGHTON} SMITH \\
& \tiny Female & \tiny \textcolor{red}{FLAT NEST} \ldots
% HAS ALSO 
\textcolor{red}{BEING} GOOD TO LAWRENCE \textcolor{red}{LADEN} SMITH \\
& \tiny Female$\rightarrow$Male & \tiny \textcolor{blue}{GLASNOST} \ldots 
% HAS ALSO 
\textcolor{red}{BIG} GOOD TO LAWRENCE \textcolor{red}{LADEN} SMITH \\

\cmidrule{2-3}
& \tiny True & \tiny BIDS \textcolor{blue}{TOTALING} SIX HUNDRED FIFTY \ldots
% ONE MILLION 
DOLLARS \ldots 
% WERE 
SUBMITTED \\
& \tiny Female & \tiny BIDS \textcolor{red}{TUMBLING} SIX HUNDRED FIFTY \ldots 
% ONE MILLION 
DOLLARS \ldots 
% WERE 
SUBMITTED \\
& \tiny Female$\rightarrow$Male & \tiny BIDS \textcolor{blue}{TOTALING} SIX HUNDRED FIFTY \ldots 
% ONE MILLION 
DOLLARS \ldots 
% WERE 
SUBMITTED \\

\cmidrule{2-3}
& \tiny True & \tiny \textcolor{blue}{NO} FIRM PLAN HAS BEEN \textcolor{blue}{DEVISED} BUT IT IS UNDER \ldots 
% CONSIDERATION TO REVIEW THE WHOLE STRUCTURE 
HE SAID \\
& \tiny Female & \tiny \textcolor{red}{NOW} FIRM PLAN HAS BEEN \textcolor{red}{DEVICES} THAT IT IS UNDER \ldots  
% CONSIDERATION TO REVIEW THE WHOLE STRUCTURE 
HE SAID \\
& \tiny Female$\rightarrow$Male & \tiny \textcolor{blue}{NO} FIRM PLAN HAS BEEN \textcolor{blue}{DEVISED} BUT IT IS UNDER \ldots 
% CONSIDERATION TO REVIEW THE WHOLE STRUCTURE 
HE SAID \\

\cmidrule{2-3}
& \tiny True & \tiny \textcolor{blue}{ALTHOUGH} SUCH EFFORTS TRIGGERED A RASH \ldots 
% OF UNSUCCESSFUL STRIKES LAST SUMMER  MANAGEMENT'S ADOPTION OF THE NEW AGGRESSIVE POSTURE HAD TO BE DONE SAYS ONE ANALYST 
\\
& \tiny Female & \tiny \textcolor{red}{ALL THIS} SUCH EFFORTS TRIGGERED A RASH \ldots 
% OF UNSUCCESSFUL STRIKES LAST SUMMER MANAGEMENT'S ADOPTION OF THE NEW AGGRESSIVE POSTURE HAD TO BE DONE SAYS ONE ANALYST 
\\
& \tiny Female$\rightarrow$Male & \tiny \textcolor{blue}{ALTHOUGH} SUCH EFFORTS TRIGGERED A RASH \ldots 
% OF UNSUCCESSFUL STRIKES LAST SUMMER MANAGEMENT'S ADOPTION OF THE NEW AGGRESSIVE POSTURE HAD TO BE DONE SAYS ONE ANALYST 
\\

\cmidrule{2-3}
& \tiny True & \tiny \textcolor{blue}{N A S A} SCHEDULED THE \textcolor{blue}{LAUNCH OF THE SPACE} SHUTTLE DISCOVERY \textcolor{blue}{FOR SEPTEMBER TWENTY NINTH} \\
& \tiny Female & \tiny \textcolor{red}{AN A F A} SCHEDULED THE \textcolor{red}{LINE ON THE STATE} SHUTTLE DISCOVERY \textcolor{red}{PRESSER} \textcolor{blue}{TWENTY NINTH} \\
& \tiny Female$\rightarrow$Male & \tiny \textcolor{red}{AN A F A} SCHEDULED THE \textcolor{blue}{LAUNCH OF THE SPACE} SHUTTLE DISCOVERY \textcolor{red}{PERSPECTIVE} \textcolor{blue}{TWENTY NINTH} \\

\cmidrule{2-3}
& \tiny True & \tiny IT MAY BE \textcolor{blue}{THAT OUR STRUCTURE} AND \ldots 
% MULTIPLICITY OF CORPORATIONS ISN'T THE MOST EFFECTIVE 
FOR THE FUTURE \\
& \tiny Female & \tiny IT MAY BE \textcolor{red}{THE ARISTECH SURE} AND \ldots
% MULTIPLICITY OF CORPORATIONS ISN'T THE MOST EFFECTIVE 
FOR THE FUTURE \\
& \tiny Female$\rightarrow$Male & \tiny IT MAY BE \textcolor{blue}{THAT OUR STRUCTURE} AND \ldots
% MULTIPLICITY OF CORPORATIONS ISN'T THE MOST EFFECTIVE 
FOR THE FUTURE \\

% \cmidrule{2-3}
% & \tiny True & \tiny \textcolor{blue}{CIBA} AGREED TO REMEDY THE \textcolor{blue}{OVERSIGHT} \\
% & \tiny Female & \tiny \textcolor{red}{SEVEN} AGREED TO REMEDY THE \textcolor{red}{OVER SITE} \\
% & \tiny Female$\rightarrow$Male & \tiny \textcolor{blue}{CIBA} AGREED TO REMEDY THE \textcolor{red}{OVER SITE} \\

% \cmidrule{2-3}
% & \tiny True & \tiny BUT LOOK A LITTLE \ldots  ARE WAYS AROUND HIRING \textcolor{blue}{FREEZES} \\
% & \tiny Female & \tiny BUT LOOK A LITTLE \ldots  ARE WAYS AROUND HIRING \textcolor{red}{PRICES} \\
% & \tiny Female$\rightarrow$Male & \tiny BUT LOOK A LITTLE \ldots  ARE WAYS AROUND HIRING \textcolor{blue}{FREEZES} \\

\bottomrule
\end{tabular}
% \vspace{-1.5em}
\end{table}

\renewcommand{\arraystretch}{1}
\begin{table}[tbp!]
\caption{ASR prediction mismatch when train on male and evaluated on female (WSJ dataset).}
  \label{tab:wsj-asr-male-2}
  \vspace{-1em}
\centering
\begin{tabular}{p{0.65cm}p{0.85cm}p{5.5cm}}
\toprule
 &  & \scriptsize Train on Male  \\
\midrule
\multirow{6}{1cm}{\scriptsize Test on Female} & \tiny True & \tiny A LITTLE GOOD NEWS COULD \textcolor{blue}{SOFTEN} THE MARKET'S RESISTANCE \\
& \tiny Female & \tiny A LITTLE GOOD NEWS COULD \textcolor{red}{SOUTH IN} THE MARKETS RESISTANCE \\
& \tiny Female$\rightarrow$Male & \tiny  A LITTLE GOOD NEWS COULD \textcolor{blue}{SOFTEN} THE MARKET'S RESISTANCE \\

\cmidrule{2-3}
& \tiny True & \tiny \textcolor{blue}{MUSICIANS ARE} MUSICIANS \\
& \tiny Female & \tiny \textcolor{red}{DESIGNS} \textcolor{blue}{ARE} MUSICIANS \\
& \tiny Female$\rightarrow$Male & \tiny \textcolor{blue}{MUSICIANS} \textcolor{red}{OR} MUSICIANS \\

\cmidrule{2-3}
& \tiny True & \tiny EARLY LAST WEEK MR \textcolor{blue}{CHUN} DID OFFER CONCESSIONS \\
& \tiny Female & \tiny EARLY LAST WEEK MR \textcolor{red}{THAN} DID OFFER CONCESSIONS \\
& \tiny Female$\rightarrow$Male & \tiny EARLY LAST WEEK MR \textcolor{blue}{CHUN} THE OFFER CONCESSIONS \\

\cmidrule{2-3}
& \tiny True & \tiny \textcolor{blue}{WE FELT THIS WAS AN ACT OF AGGRESSION HE SAID WITHOUT ANY MORAL OR POLITICAL JUSTIFICATION} \\
& \tiny Female & \tiny \textcolor{blue}{WE} \textcolor{red}{THOUGHT THIS LIES AN ACTIVE} \textcolor{blue}{AGGRESSION} \textcolor{blue}{HE SAID} \textcolor{red}{WITH THAT} \textcolor{blue}{ANY MORAL OR POLITICAL} \textcolor{red}{DESTINATION} \\
& \tiny Female$\rightarrow$Male & \tiny \textcolor{blue}{WE FELT THIS WAS AN} \textcolor{red}{ACTIVE} \textcolor{blue}{AGGRESSION} \textcolor{blue}{HE SAID WITHOUT ANY MORE ALL OUR POLITICAL JUSTIFICATION} \\

\cmidrule{2-3}
& \tiny True & \tiny \textcolor{blue}{THE REMAINING} NINETY NINE PERCENT INTEREST IN THE PARTNERSHIP CURRENTLY IS HELD BY \textcolor{blue}{AFFILIATES OF AIRCOA} \\
& \tiny Female & \tiny \textcolor{red}{THEY REMAIN IN} NINETY NINE PERCENT INTEREST IN THE PARTNERSHIP CURRENTLY IS HELD BY \textcolor{red}{AFFILIATE TO THEIR COLA} \\
& \tiny Female$\rightarrow$Male & \tiny \textcolor{blue}{THE REMAINING} NINETY NINE PERCENT INTEREST IN THE PARTNERSHIP CURRENTLY IS HELD BY \textcolor{blue}{AFFILIATES OF THE AIRCOA} \\

\cmidrule{2-3}
& \tiny True & \tiny \textcolor{blue}{BUT EVEN THIS SILVER HAIRED CIGAR SMOKING DIPLOMAT USED TOUGH WORDS TO DESCRIBE AMERICA'S ARMS SALES TO IRAN} \\
& \tiny Female & \tiny \textcolor{blue}{BUT EVEN} \textcolor{red}{THE} \textcolor{blue}{SILVER} \textcolor{red}{HAD A GRASPING} \textcolor{blue}{DIPLOMAT} \textcolor{red}{EAST} \textcolor{blue}{TOUGH WORDS TO DESCRIBE AMERICA'S ARMS SALES} \textcolor{red}{TAIWAN} \\
& \tiny Female$\rightarrow$Male & \tiny \textcolor{blue}{BUT EVEN THIS SILVER HAIRED} \textcolor{red}{SUGAR} \textcolor{blue}{SMOKING DIPLOMAT} \textcolor{red}{USE} \textcolor{blue}{TOUGH WORTH TO DESCRIBE AMERICA'S ARMS SALES TO IRAN} \\

\cmidrule{2-3}
& \tiny True & \tiny \textcolor{blue}{GEORGE E R KINNEAR} THE SECOND WAS NAMED \textcolor{blue}{TO} THE NEW POST OF SENIOR VICE PRESIDENT IN CHARGE OF LONG RANGE PLANNING \textcolor{blue}{AND} RELATED GOVERNMENT RELATIONS \\
& \tiny Female & \tiny \textcolor{red}{JORGE E} \textcolor{blue}{KINNEAR} THE SECOND WAS NAMED \textcolor{red}{DID} THE NEW POST OF SENIOR VICE PRESIDENT IN CHARGE OF LONG RANGE PLANNING \textcolor{red}{IN} RELATED GOVERNMENT RELATIONS \\
& \tiny Female$\rightarrow$Male & \tiny \textcolor{blue}{GEORGE E} \textcolor{red}{I} \textcolor{blue}{KINNEAR} THE SECOND WAS NAMED \textcolor{blue}{TO} THE NEW POST OF SENIOR VICE PRESIDENT IN CHARGE OF LONG RANGE PLANNING \textcolor{red}{IN} RELATED GOVERNMENT RELATIONS \\

\cmidrule{2-3}
& \tiny True & \tiny HE ARGUES THAT \textcolor{blue}{FRIDAY'S UNEMPLOYMENT} FIGURES UNDERMINED \textcolor{blue}{THE THESIS OF} A SHARPLY SLOWING ECONOMY \\
& \tiny Female & \tiny HE ARGUES THAT \textcolor{blue}{FRIDAY'S} \textcolor{red}{AND EMPLOYMENT} FIGURES UNDERMINED \textcolor{red}{THAT THESE SAYS THEM} A SHARPLY SLOWING ECONOMY \\
& \tiny Female$\rightarrow$Male & \tiny HE ARGUES THAT \textcolor{blue}{FRIDAY'S UNEMPLOYMENT} FIGURES UNDERMINED \textcolor{blue}{THE THESIS OF} A SHARPLY SLOWING ECONOMY \\

\cmidrule{2-3}
& \tiny True & \tiny THERE ARE SOME \textcolor{blue}{GUYS HERE} THAT ARE \textcolor{blue}{SAYING} THIS IS THE FINAL JUMP BEFORE THE CRASH \\
& \tiny Female & \tiny THERE ARE SOME \textcolor{red}{DIES YEAR} THAT ARE \textcolor{blue}{SAYING} THIS IS THE FINAL JUMP BEFORE THE CRASH \\
& \tiny Female$\rightarrow$Male & \tiny THERE ARE SOME \textcolor{blue}{GUYS} \textcolor{red}{YEAR} THAT ARE \textcolor{red}{SEEING} THIS IS THE FINAL JUMP BEFORE THE CRASH \\

\cmidrule{2-3}
& \tiny True & \tiny ONLY \textcolor{blue}{A FEW} STATES REQUIRE UNEMPLOYMENT COMPENSATION FOR LOCKED OUT WORKERS \\
& \tiny Female & \tiny ONLY \textcolor{red}{IF HE'S} STATES REQUIRE UNEMPLOYMENT COMPENSATION FOR LOCKED OUT WORKERS \\
& \tiny Female$\rightarrow$Male & \tiny ONLY \textcolor{blue}{A FEW} STATES REQUIRE UNEMPLOYMENT COMPENSATION FOR LOCKED OUT WORKERS \\

\cmidrule{2-3}
& \tiny True & \tiny HE ALSO SAYS THE AUTHORITIES MEAN WHAT THEY SAY THAT THEY WILL NOT STAND ASIDE AND \textcolor{blue}{LET CURRENCIES REACH} NEW LOWS POST ELECTION \\
& \tiny Female & \tiny HE ALSO SAYS THE AUTHORITIES MEAN WHAT THEY SAY THAT THEY WILL NOT STAND ASIDE AND \textcolor{red}{LIGHT FRANCE'S REACHED} NEW LOWS POST ELECTION \\
& \tiny Female$\rightarrow$Male & \tiny HE ALSO SAYS THE AUTHORITIES MEAN WHAT THEY SAY THAT THEY WILL NOT STAND ASIDE AND \textcolor{blue}{LET CURRENCIES REACH} NEW LOWS POST ELECTION \\

\cmidrule{2-3}
& \tiny True & \tiny BUT LOOK A LITTLE FURTHER THERE ARE WAYS AROUND HIRING \textcolor{blue}{FREEZES} \\
& \tiny Female & \tiny BUT LOOK A LITTLE FURTHER THERE ARE WAYS AROUND HIRING \textcolor{red}{PRICES} \\
& \tiny Female$\rightarrow$Male & \tiny BUT LOOK A LITTLE FURTHER THERE ARE WAYS AROUND HIRING \textcolor{blue}{FREEZES} \\

\cmidrule{2-3}
& \tiny True & \tiny MR HOLMES \textcolor{blue}{A COURT} SAID HE PLANS TO REVIEW THE STRUCTURE OF \textcolor{blue}{BELL GROUP} \\
& \tiny Female & \tiny MR HOLMES \textcolor{red}{ACQUIRED} SAID HE PLANS TO REVIEW THE STRUCTURE OF \textcolor{red}{BALLET} \\
& \tiny Female$\rightarrow$Male & \tiny MR HOLMES \textcolor{blue}{A COURT} SAID HE PLANS TO REVIEW THE STRUCTURE OF \textcolor{blue}{BELL GROUP} \\

\bottomrule

\end{tabular}
\end{table}

\begin{table}[tbp!]
\caption{ASR prediction mismatch when train on female and evaluated on male (WSJ dataset).}
  \label{tab:wsj-asr-female-1}
  \vspace{-1em}
\centering
\begin{tabular}{p{0.65cm}p{0.85cm}p{5.5cm}}

\toprule
 &  & \scriptsize Train on Female  \\
\midrule
\multirow{6}{1cm}{\scriptsize Test on Male} & \tiny True & \tiny STRONGER PALM \textcolor{blue}{OIL} PRICES HELPED OIL \textcolor{blue}{PRICES FIRM} ANALYSTS \textcolor{blue}{SAID} \\
& \tiny Male & \tiny STRONGER PALM \textcolor{red}{ON} PRICES HELPED OIL \textcolor{red}{PRICE FROM} ANALYSTS \textcolor{red}{SO} \\
& \tiny Male$\rightarrow$Female & \tiny STRONGER PALM \textcolor{blue}{OIL} PRICES HELPED OIL \textcolor{blue}{PRICES FIRM} ANALYSTS \textcolor{blue}{SAID} \\

\cmidrule{2-3}
& \tiny True & \tiny CONTACTS STILL INSIDE OWENS \textcolor{blue}{CORNING} HELP \textcolor{blue}{TOO} \\
& \tiny Male & \tiny CONTACTS STILL INSIDE OWENS \textcolor{red}{CORN AND} HELP \textcolor{red}{TO} \\
& \tiny Male$\rightarrow$Female & \tiny CONTACTS STILL INSIDE OWENS \textcolor{blue}{CORNING} HELP \textcolor{red}{TO} \\

\cmidrule{2-3}
& \tiny True & \tiny THE \textcolor{blue}{WARMING TREND MAY HAVE} MELTED THE SNOW COVER ON SOME CROPS \\
& \tiny Male & \tiny THE \textcolor{red}{WOMAN} \textcolor{blue}{TREND} \textcolor{red}{MAYOR} MELTED THE SNOW COVER ON SOME CROPS \\
& \tiny Male$\rightarrow$Female & \tiny THE \textcolor{blue}{WARMING TREND MAY} \textcolor{red}{HAD} MELTED TO SNOW COVER ON SOME CROPS \\

\cmidrule{2-3}
& \tiny True & \tiny HE MADE A SALES CALL HE \textcolor{blue}{SAYS} \\
& \tiny Male & \tiny HE MADE A SALES CALL HE \textcolor{red}{SERVES} \\
& \tiny Male$\rightarrow$Female & \tiny HE MADE A SALES CALL HE \textcolor{blue}{SAYS} \\

\cmidrule{2-3}
& \tiny True & \tiny IT WASN'T \textcolor{blue}{A GIVEAWAY} \\
& \tiny Male & \tiny IT WASN'T \textcolor{red}{TO GIVE MORE} \\
& \tiny Male$\rightarrow$Female & \tiny IT WASN'T \textcolor{red}{THAT GIVE AWAY} \\

\cmidrule{2-3}
& \tiny True & \tiny VICE PRESIDENT BUSH MUST BE ESPECIALLY \textcolor{blue}{GRATEFUL} FOR THE CHANGE OF SUBJECT  ANYTHING WAS BETTER THAN THE DRUMBEAT ABOUT PANAMA AND \textcolor{blue}{GENERAL NORIEGA} \\
& \tiny Male & \tiny VICE PRESIDENT BUSH MUST BE ESPECIALLY \textcolor{red}{GREAT FULL} FOR THE CHANGE OF SUBJECT ANYTHING WAS BETTER THAN \textcolor{red}{A} DRUMBEAT ABOUT PANAMA AND \textcolor{red}{TRUMAN MILEAGE} \\
& \tiny Male$\rightarrow$Female & \tiny VICE PRESIDENT BUSH MUST BE ESPECIALLY \textcolor{red}{GREAT FULL} FOR THE CHANGE OF SUBJECT ANYTHING WAS BETTER THAN THE DRUM BEAT ABOUT PANAMA AND \textcolor{blue}{GENERAL NORIEGA} \\

\cmidrule{2-3}
& \tiny True & \tiny IN NINETEEN EIGHTY FIVE PENNZOIL \textcolor{blue}{WON} NEARLY ELEVEN BILLION DOLLARS IN DAMAGES \textcolor{blue}{AT TRIAL THE BIGGEST JUDGMENT EVER AWARDED} A PLAINTIFF \\
& \tiny Male & \tiny IN NINETEEN EIGHTY FIVE PENNZOIL \textcolor{red}{ONE} NEARLY ELEVEN BILLION DOLLARS IN DAMAGES \textcolor{red}{ARE} \textcolor{blue}{THE BIGGEST} \textcolor{red}{GEORGE MEN EVEN ORDERED} A PLAINTIFF \\
& \tiny Male$\rightarrow$Female & \tiny IN NINETEEN EIGHTY FIVE PENNZOIL \textcolor{red}{ONE} NEARLY ELEVEN BILLION DOLLARS IN DAMAGES \textcolor{red}{AT TRY} \textcolor{blue}{THE BIGGEST JUDGMENT EVER AWARDED} A PLAINTIFF \\

\cmidrule{2-3}
& \tiny True & \tiny BUT IT'S DIFFICULT TO SEE WHERE THE COMPANY \textcolor{blue}{GOES FROM HERE} \\
& \tiny Male & \tiny BUT IT'S DIFFICULT TO SEE WHERE THE COMPANY \textcolor{red}{GOT YEAR} \\
& \tiny Male$\rightarrow$Female & \tiny BUT IT'S DIFFICULT TO SEE WHERE THE COMPANY \textcolor{blue}{GOES FROM HERE} \\

\cmidrule{2-3}
& \tiny True & \tiny \textcolor{blue}{THEY EXPECT} COMPANIES \textcolor{blue}{TO GROW OR} DISAPPEAR \\
& \tiny Male & \tiny \textcolor{red}{THE DEBUT} COMPANIES TO \textcolor{red}{GO ON} DISAPPEAR \\
& \tiny Male$\rightarrow$Female & \tiny \textcolor{blue}{THEY EXPECT} COMPANIES \textcolor{blue}{TO GROW OR} DISAPPEAR \\

\cmidrule{2-3}
& \tiny True & \tiny ALSO \textcolor{blue}{MENTIONED WAS A CONTROVERSIAL PROPOSAL} TO DENY THE DEDUCTION FOR TWENTY PERCENT OF CORPORATE ADVERTISING COSTS \textcolor{blue}{AND TO REQUIRE INSTEAD THAT THEY BE AMORTIZED OVER TWO YEARS} \\
& \tiny Male & \tiny ALSO \textcolor{red}{NOTION WERE RETREAT} \textcolor{blue}{PROPOSAL} TO DENY THE DEDUCTION FOR TWENTY PERCENT OF CORPORATE ADVERTISING COSTS \textcolor{blue}{AND} \textcolor{red}{REQUIRING STOP BUT THE} \textcolor{blue}{BE AMORTIZED OVER} \textcolor{red}{TOURS} \\
& \tiny Male$\rightarrow$Female & \tiny ALSO \textcolor{blue}{MENTIONED WAS A CONTROVERSIAL PROPOSAL} TO DENY THE DEDUCTION FOR A TWENTY PERCENT OF CORPORATE ADVERTISING COSTS \textcolor{blue}{AND TO REQUIRE INSTEAD THAT THEY BE AMORTIZED OVER TO YEARS} \\

\cmidrule{2-3}
& \tiny True & \tiny ALTHOUGH JAPAN'S POLICIES WON'T CHANGE RADICALLY \textcolor{blue}{UNDER} TAKESHITA SIXTY \textcolor{blue}{THREE} HIS LACK OF FOREIGN POLICY EXPERIENCE COULD \textcolor{blue}{WORSEN JAPAN'S} INTERNATIONAL RELATIONS \\
& \tiny Male & \tiny ALTHOUGH JAPAN'S POLICIES WON'T CHANGE RADICALLY \textcolor{red}{ON THE} TAKESHITA SIXTY \textcolor{red}{FOR} HIS LACK OF FOREIGN POLICY EXPERIENCE COULD \textcolor{blue}{WORSEN} \textcolor{red}{REPAIRS} INTERNATIONAL RELATIONS \\
& \tiny Male$\rightarrow$Female & \tiny ALTHOUGH JAPAN'S POLICIES WON'T CHANGE RADICALLY \textcolor{blue}{UNDER} TAKESHITA SIXTY \textcolor{blue}{THREE} HAS LACK OF FOREIGN POLICY EXPERIENCE COULD \textcolor{red}{WORSE IN} \textcolor{blue}{JAPAN'S} INTERNATIONAL RELATIONS \\

\cmidrule{2-3}
& \tiny True & \tiny HOWEVER INCREASING THE COST OF RESEARCH ANIMALS SHOULD \textcolor{blue}{MOTIVATE} RESEARCHERS NOT TO WASTE THEM ON \textcolor{blue}{MERELY} CURIOUS OR \textcolor{blue}{REPETITIVE} STUDIES \\
& \tiny Male & \tiny HOWEVER INCREASING THE COST OF RESEARCH ANIMALS SHOULD \textcolor{red}{MOTIVE AT} RESEARCHERS NOT TO WASTE THEM ON \textcolor{red}{NEARLY} CURIOUS OR \textcolor{red}{REPEATED} STUDIES \\
& \tiny Male$\rightarrow$Female & \tiny HOWEVER INCREASING THE COST OF RESEARCH ANIMALS SHOULD \textcolor{blue}{MOTIVATE} RESEARCHERS NOT TO WASTE THEM ON \textcolor{blue}{MERELY} CURIOUS OR \textcolor{red}{REPEATED OF} STUDIES \\

\bottomrule
\end{tabular}
% \vspace{-1.5em}
\end{table}

\begin{table}[tbp!]
\caption{ASR prediction mismatch when train on female and evaluated on male (WSJ dataset).}
  \label{tab:wsj-asr-female-2}
  \vspace{-1em}
\centering
\begin{tabular}{p{0.65cm}p{0.85cm}p{5.5cm}}

\toprule
 &  & \scriptsize Train on Female  \\
\midrule
\multirow{6}{1cm}{\scriptsize Test on Male} & \tiny True & \tiny \textcolor{blue}{AND} SOME CHAINS SUCH AS HOLIDAY CORPORATION \textcolor{blue}{SHERATON} CORPORATION \textcolor{blue}{AND HYATT HOTELS} CORPORATION INSIST THEY WILL MAKE \textcolor{blue}{PLENTY} OF ROOMS AVAILABLE \textcolor{blue}{AT BARGAIN RATES} \\
& \tiny Male & \tiny \textcolor{red}{IN} SOME CHAINS SUCH AS HOLIDAY CORPORATION \textcolor{red}{SHARES} CORPORATION \textcolor{red}{TERMINATE LES} CORPORATION INSIST THEY WILL MAKE \textcolor{red}{PLAIN} OF ROOMS AVAILABLE \textcolor{red}{OR} \textcolor{blue}{BARGAIN} \textcolor{red}{ROUTES} \\
& \tiny Male$\rightarrow$Female & \tiny \textcolor{blue}{AND} SOME CHAINS SUCH AS HOLIDAY CORPORATION \textcolor{red}{SHARES IN} CORPORATION \textcolor{blue}{AND} \textcolor{red}{HIGH AT} HOTELS CORPORATION INSIST THEY WILL MAKE \textcolor{blue}{PLENTY} OF ROOMS AVAILABLE \textcolor{blue}{AT BARGAIN RATES} \\

\cmidrule{2-3}
& \tiny True & \tiny THE NASDAQ COMPOSITE \textcolor{blue}{INDEX OF} FOUR THOUSAND SIX HUNDRED THIRTY EIGHT STOCKS CLOSED AT THREE HUNDRED SEVENTY FOUR POINT SIX FIVE DOWN ZERO POINT ONE \textcolor{blue}{SIX} \\
& \tiny Male & \tiny THE NASDAQ COMPOSITE \textcolor{red}{IN BARS AND} FOUR THOUSAND SIX HUNDRED THIRTY EIGHT STOCKS CLOSED AT THREE HUNDRED SEVENTY FOUR POINT SIX FIVE DOWN ZERO POINT ONE \textcolor{red}{SOUTH} \\
& \tiny Male$\rightarrow$Female & \tiny THE NASDAQ COMPOSITE \textcolor{blue}{INDEX OF} FOUR THOUSAND SIX HUNDRED THIRTY EIGHT STOCKS CLOSED AT THREE HUNDRED SEVENTY FOUR POINT SIX FIVE DOWN ZERO POINT ONE \textcolor{blue}{SIX} \\

\cmidrule{2-3}
& \tiny True & \tiny ALTHOUGH MOST OF \textcolor{blue}{THOSE} HAVE BEEN WEAK PERFORMERS \textcolor{blue}{THIS YEAR} THAT HASN'T STOPPED OTHERS FROM TRYING TO \textcolor{blue}{CASH IN ON THE TERM'S NEW CACHET} \\
& \tiny Male & \tiny ALTHOUGH MOST OF \textcolor{red}{LOANS} HAVE BEEN WEAK PERFORMERS \textcolor{red}{LOSER} THAT HASN'T STOPPED OTHERS FROM TRYING TO \textcolor{blue}{CASH} \textcolor{red}{ON} \textcolor{blue}{THE} \textcolor{red}{TERMS ON CASH} \\
& \tiny Male$\rightarrow$Female & \tiny ALTHOUGH MOST OF \textcolor{blue}{THOSE} HAVE BEEN WEAK PERFORMERS \textcolor{blue}{THIS YEAR} THAT HASN'T STOPPED OTHERS FROM TRYING TO \textcolor{blue}{CASH IN ON THE} \textcolor{blue}{TERM'S NEW} \textcolor{red}{CASH} \\

\cmidrule{2-3}
& \tiny True & \tiny MARKET ACTION WAS ESSENTIALLY \textcolor{blue}{DIGESTING WEDNESDAY'S RALLY} \\
& \tiny Male & \tiny MARKET ACTION WAS ESSENTIALLY \textcolor{red}{BY JUST IN ONE DAY'S RODE} \\
& \tiny Male$\rightarrow$Female & \tiny MARKET \textcolor{red}{IN} ACTION WAS ESSENTIALLY \textcolor{red}{IN} \textcolor{blue}{DIGESTING WEDNESDAY'S RALLY} \\

\cmidrule{2-3}
& \tiny True & \tiny NUCLEAR WEAPONS \textcolor{blue}{FREE ZONES} ARE ATTRACTING INCREASING \textcolor{blue}{AND MISGUIDED PUBLIC AND PARLIAMENTARY} ATTENTION IN THE POST \textcolor{blue}{I N F} WESTERN WORLD \\
& \tiny Male & \tiny NUCLEAR WEAPONS \textcolor{red}{FREEDOMS} ARE ATTRACTING INCREASING \textcolor{red}{UNDERSCORED} \textcolor{blue}{PUBLIC} \textcolor{red}{IN PARLIAMENT ARE} ATTENTION IN THE POST \textcolor{red}{I M F} WESTERN WORLD \\
& \tiny Male$\rightarrow$Female & \tiny NUCLEAR WEAPONS \textcolor{red}{FREEZES} ARE ATTRACTING INCREASING \textcolor{blue}{AT MISGUIDED PUBLIC AND PARLIAMENTARY} ATTENTION IN THE POST \textcolor{blue}{I N F} WESTERN WORLD \\

\cmidrule{2-3}
& \tiny True & \tiny THE RECALL EXPANDS ON A \textcolor{blue}{WITHDRAWAL OF OTHER MODELS BEGUN EARLIER THIS WEEK} \\
& \tiny Male & \tiny THE RECALL EXPANDS ON A \textcolor{red}{WERE WELL OFF OVER} \textcolor{blue}{MODELS} \textcolor{red}{BOONE ROLE OF IS QUICK} \\
& \tiny Male$\rightarrow$Female & \tiny THE RECALL EXPANDS ON A \textcolor{red}{WAS DRAW ALL} \textcolor{blue}{OF OTHER MODELS} \textcolor{red}{BEGAN} \textcolor{blue}{EARLIER THIS WEEK} \\

\cmidrule{2-3}
& \tiny True & \tiny THRIFT \textcolor{blue}{NET WORTH} \\
& \tiny Male & \tiny THRIFT \textcolor{red}{NATWEST} \\
& \tiny Male$\rightarrow$Female & \tiny THRIFT \textcolor{blue}{NET} \textcolor{red}{WORSE} \\

\cmidrule{2-3}
& \tiny True & \tiny MR \textcolor{blue}{POLO} ALSO \textcolor{blue}{OWNS} THE FASHION COMPANY \\
& \tiny Male & \tiny MR \textcolor{red}{PAYING} ALSO \textcolor{red}{LONG} THE FASHION COMPANY \\
& \tiny Male$\rightarrow$Female & \tiny MR \textcolor{blue}{POLO} ALSO \textcolor{blue}{OWNS} THE FASHION COMPANY \\

\cmidrule{2-3}
& \tiny True & \tiny THE FATE OF \textcolor{blue}{THE UNIVERSE} IS STILL A MYSTERY \\
& \tiny Male & \tiny THE FATE OF \textcolor{red}{VIRUS} IS STILL A MYSTERY \\
& \tiny Male$\rightarrow$Female & \tiny THE FATE OF \textcolor{blue}{THE UNIVERSE} IS STILL A MYSTERY \\

\cmidrule{2-3}
& \tiny True & \tiny \textcolor{blue}{ASTRONOMERS SAY THAT THE EARTH'S FATE IS SEALED} \\
& \tiny Male & \tiny \textcolor{red}{TRADERS} \textcolor{blue}{SAY} \textcolor{red}{OF THE FAT IS SOLD} \\
& \tiny Male$\rightarrow$Female & \tiny \textcolor{blue}{ASTRONOMERS SAY THAT THE EARTH'S FATE IS} \textcolor{red}{SHIELD} \\

\cmidrule{2-3}
& \tiny True & \tiny \textcolor{blue}{FIVE BILLION YEARS FROM NOW THE SUN WILL SLOWLY SWALLOW THE EARTH IN A HUGE FIREBALL} \\
& \tiny Male & \tiny \textcolor{red}{FAR} \textcolor{blue}{BILLION YEARS} \textcolor{red}{SHUNNED THE SHOUTING SLOWER SWELL LEON SOARED} \textcolor{blue}{A HUGE} \textcolor{red}{PARABLE} \\
& \tiny Male$\rightarrow$Female & \tiny \textcolor{blue}{FIVE BILLION YEARS} \textcolor{red}{SHADOW} \textcolor{blue}{THE SUN WILL} \textcolor{red}{SLOW ACTUALLY EARNS ON} \textcolor{blue}{A HUGE FIREBALL} \\

\cmidrule{2-3}
& \tiny True & \tiny \textcolor{blue}{DRAVO} LAST MONTH \textcolor{blue}{AGREED} IN PRINCIPLE TO SELL ITS \textcolor{blue}{INLAND} WATER TRANSPORTATION \textcolor{blue}{STEVEDORING} AND PIPE FABRICATION BUSINESSES FOR AN UNDISCLOSED SUM \\
& \tiny Male & \tiny \textcolor{red}{TRAVEL} LAST MONTH \textcolor{red}{FOR GREED} IN PRINCIPLE TO SELL ITS \textcolor{red}{IMPLIED} WATER TRANSPORTATION \textcolor{red}{STARRING} AND PIPE FABRICATION BUSINESSES FOR AN UNDISCLOSED SUM \\
& \tiny Male$\rightarrow$Female & \tiny \textcolor{red}{TRAVEL} LAST MONTH \textcolor{blue}{AGREED} IN PRINCIPLE TO SELL ITS \textcolor{blue}{INLAND} WATER TRANSPORTATION \textcolor{blue}{STEVEDORING} AND PIPE FABRICATION BUSINESSES FOR AN UNDISCLOSED SUM \\

\cmidrule{2-3}
& \tiny True & \tiny \textcolor{blue}{UNFORTUNATELY WE'VE SEEN NOTHING BUT STUNTS} \\
& \tiny Male & \tiny \textcolor{red}{ON FORTUNE ALL WE'RE SEEING} \textcolor{blue}{NOTHING BUT} \textcolor{red}{STOCKS} \\
& \tiny Male$\rightarrow$Female & \tiny \textcolor{red}{AND FORTUNATELY} \textcolor{blue}{WE'VE SEEN NOTHING BUT} \textcolor{blue}{STUNTS} \\

\bottomrule
\end{tabular}
\end{table}
  
\end{appendices}

\end{document}